\documentclass[10pt,twocolumn,letterpaper]{article}

\usepackage{wacv}
\usepackage{times}
\usepackage{epsfig}
\usepackage{graphicx}
\usepackage{amsmath}
\usepackage{amssymb}

\usepackage{color}
\usepackage{commath}
\usepackage{multirow}
\usepackage{wrapfig}
\usepackage{array}
\usepackage{booktabs}

\def\wacvPaperID{712} 

\wacvfinalcopy 
\ifwacvfinal
\def\assignedStartPage{9876} 
\fi

\ifwacvfinal
\usepackage[breaklinks=true,bookmarks=false]{hyperref}
\else
\usepackage[pagebackref=true,breaklinks=true,colorlinks,bookmarks=false]{hyperref}
\fi

\ifwacvfinal
\else
\fi

\begin{document}

\title{Improve CAM with Auto-adapted Segmentation and Co-supervised Augmentation}

\author{
Ziyi Kou\thanks{This work was done while they were at University of Rochester.} $^{\ \dagger}$\\
University of Notre Dame\\
{\tt\small zkou@nd.edu}
\and
$\text{Guofeng Cui}^\text{*}$ \thanks{Both authors contributed equally to this work.} \\
Rutgers University\\
{\tt\small gc669@cs.rutgers.edu}
\and
$\text{Shaojie Wang}^\text{*}$\\ 
Washington University in St. Louis\\
{\tt\small joss@wustl.edu}
\and
$\text{Wentian Zhao}^\text{*}$\\
Adobe\\
{\tt\small wezhao@adobe.com}
\and
Chenliang Xu\\
University of Rochester\\
{\tt\small chenliang.xu@rochester.edu}
}

\maketitle

\begin{abstract}
Weakly Supervised Object Localization (WSOL) methods generate both classification and localization results by learning from only image category labels. Previous methods usually utilize class activation map (CAM) to obtain target object regions. However, most of them only focus on improving foreground object parts in CAM, but ignore the important effect of its background contents. In this paper, we propose a confidence segmentation (ConfSeg) module that builds confidence score for each pixel in CAM without introducing additional hyper-parameters. The generated sample-specific confidence mask is able to indicate the extent of determination for each pixel in CAM, and further supervises additional CAM extended from internal feature maps. Besides, we introduce Co-supervised Augmentation (CoAug) module to capture feature-level representation for foreground and background parts in CAM separately. Then a metric loss is applied at batch sample level to augment distinguish ability of our model, which helps a lot to localize more related object parts. Our final model, CSoA, combines the two modules and achieves superior performance, e.g. $37.69\%$ and $48.81\%$ Top-1 localization error on CUB-200 and ILSVRC datasets, respectively, which outperforms all previous methods and becomes the new state-of-the-art.
\end{abstract}

\section{Introduction}

Weakly-Supervised Object Localization (WSOL) aims to learn object locations in a given image from only image-level labels. It avoids expensive bounding box annotations and thus dramatically reduces the cost of human labors in image annotations. To tackle the problem, utilizing class activation map (CAM) is often adopted as a good choice recently. CAM is a type of 3D feature map with each channel corresponding to one category label. The pixels in it can indicate the discriminative regions of objects belonging to each category. Therefore, by extracting the features via the label index, the model can roughly locate the position of target objects. The main reason for the wide use of CAM is that the generation of CAM needs only little modifications based on classical CNN backbones but the performance is robust. For instance, Zhou et al.~\cite{Zhou_2016_CVPR} propose to replace fully connected layer with global average pooling layer (GAP) to generate CAM for a given image, which achieves a competitive localization result.

\begin{figure*}[t]
    \centering
    \includegraphics[width=1.0\linewidth]{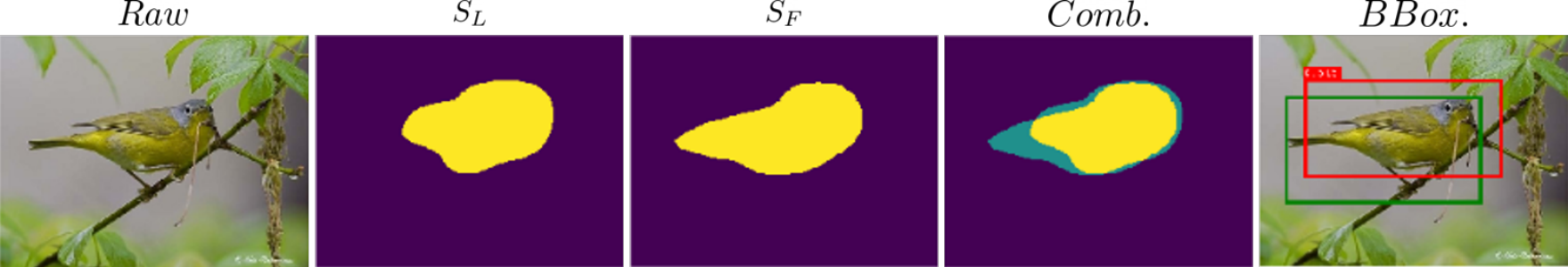}
    \caption{Binary localization maps from two CAMs and the final localization result. $S_L$ is from CAM at the top layer while $S_F$ from another one. The combination of two maps and the final bounding box demonstrates the advantage of our network to produce more complete and complementary results. Red and green bounding boxes denotes predicted localization result can ground-truth label.}
    \label{fig:ConfSeg}
\end{figure*}

Though using CAM for localization is efficient and straightforward, it can only detect some parts of the objects instead of covering the full object extents. The main reason is that traditional classification networks tend to distinguish images by focusing on the most representative regions, which can minimize the classification loss but results in losing other related but non-essential parts. To address the problem, lots of approaches~\cite{Singh_2017_ICCV,Zhang_2018_CVPR,Zhang_2018_ECCV,Choe_2019_CVPR,Xue_2019_ICCV,combine_wacv20} have been proposed, and they can be categorized roughly into the following two classes. The first class of methods~\cite{Singh_2017_ICCV,Choe_2019_CVPR} manipulates input data samples or internal feature maps directly to enforce the network to search related object parts. It improves localization but sacrifices classification performance because the target objects may become unrecognized after their parts are erased. The second type of methods~\cite{Zhang_2018_CVPR,Xue_2019_ICCV,combine_wacv20} generate multiple CAMs and combine them for the final localization. Their CAMs are useful as they contain information from different convolutional layers or different levels of semantics.

However, 
all the above methods only focus on expanding foreground object regions and ignore background parts in CAM. In our observation, determining background not only helps remove unrelated pixels but also plays as additional supervisions when multiple CAMs are being used. 
Indeed, to the best of our knowledge, \cite{Zhang_2018_ECCV} is the only work that considers segmenting background contents of CAM based on internal feature maps. However, it has to set fixed segmentation thresholds for all samples in a one-size-fit-all manner, which is not optimal. Besides, the background segmentation in \cite{Zhang_2018_ECCV} is only used to regulate a single CAM during training and discarded in the inference time. Therefore, such approach is not an ideal way to generate and utilize background regions to improve localization performance.

Though the CAM can be self-refined by internal feature maps as discussed above, no additional regularization for generated CAM is provided in previous methods. Supervised by only category-level labels, CAM becomes unstable for localization. For instance, \cite{wang2018weakly} indicates for samples belonging to the same category, the model will focus on different object parts due to the various characteristics the target object displays. However, such phenomenon is not expected in our localization task since we prefer the complete prediction of objects for each sample.


To overcome the above limitations, we propose a new framework named CSoA for the WSOL task with two novel modules. We first introduce the confidence segmentation (ConfSeg) module, an internal module that connects and refines two different CAMs inside our network. One of the two CAMs is generated from the top convolutional layer and thus captures high-level semantic information. Another CAM is extended from internal feature maps of the backbone network and contains fine-level object boundary clues. These two CAMs have totally different characteristics and receptive fields but both contribute to the final localization and classification performance. With these two different CAMs, the ConfSeg module segments a dynamic per-sample confidence mask from the first CAM and applies it as additional supervisions to regulate the second one, which finally encourages them to be incorporated together with both high-level and fine-level information. Fig. \ref{fig:ConfSeg} shows the final binary localization maps extracted from the two CAMs and their combination. Especially, the generated maps concentrate on foreground object parts with similar center area but become complementary for surrounded regions. Without introducing additional hyper-parameters, the ConfSeg module greatly improves the final localization performance compared with the current state-of-the-art results. 

Apart from the ConfSeg module, we propose a metric-based approach denoted as Co-supervised Augmentation (CoAug) regularizer to further regulate CAM and augment its integrity. For CoAug, two expectations for CAM are considered. The first one is that an ideal CAM should separate the image into two regions with non-intersected contents, specifically foreground objects and background. The second one is aligned according to foreground regions focused by CAM of different samples. For images belonging to the same category, their foreground parts are supposed to share similar identifications. While for different categories, the samples should be distinct with each other. With the above two assumptions, we construct the CoAug module that enforces batch-level samples to inter-supervise collaboratively by embedded vectors that are also applied in ~\cite{hsu2018co,koch2015siamese}. By this way, the CoAug module enhances the recognition ability of CAM by not only gathering the information of each category from various samples, but also discriminate them. The details of the module will be discussed in Section~\ref{sec:method} and its advantage will be demonstrated in Section~\ref{sec:exp}.


In summary, our main contributions are three folds: (1) We propose a novel confidence segmentation module to generate a confidence mask that gets two different CAMs interacted and refined without additional hyper parameters. (2) We propose Co-supervised Augmentation module to refine CAM by regulating feature-level representations, which guides our model to localize more related object regions. (3) With only image-level supervision for training, our method greatly outperforms other state-of-the-art methods on two standard benchmarks, ILSVRC validation set and CUB-200-2011 test set, for weakly supervised localization performance. 

\section{Related Work}

Weakly Supervised Object Localization usually relies on CAM to localize objects. Zhou et al.~\cite{Zhou_2016_CVPR} propose Global Average Pooling (GAP) layer for deep neural networks to generate CAM for localization. Based on it, Zhang et al.~\cite{Zhang_2018_CVPR} prove that the process for obtaining  CAM  can  be  end-to-end and further propose ACoL network that adopts cut-and-search strategy on the feature maps. Moreover, Zhang et al.~\cite{Zhang_2018_ECCV} propose SPG network that extends pixel-level mask from internal feature maps and complement CAM in the final. Recently, Choe et al.~\cite{Choe_2019_CVPR} design a general dropout algorithm for internal feature maps to refine CAM from bottom level. Besides, the clustering of ground-truth labels is proposed in ~\cite{Xue_2019_ICCV} to obtain multiple semantic level CAM. 

Similar to the WSOL problem, the methods for Object Co-Segmentation task attempts to segment target objects based on image-level labels. However, as introduced by Rother et al~\cite{rother2006cosegmentation}, co-segmentation task aims to segment common objects from a set of images belonging to a specific category instead of multiple ones. The main idea for tackling the problem is to leverage intra-image discovery and inter-sample correlation~\cite{hsu2018co,chen2018semantic,li2018unsupervised,li2018deep,tao2017image}. For example, Li et al.~\cite{li2018deep} embed image features by Siamese Encoder and then apply feature matching to extract common objects from image pairs. Hsu et al.~\cite{hsu2018co} introduce co-attention loss based on intra- and inter-sample comparison to guide the object discovery process. Besides, they utilize unsupervised methods to pick object proposals in order to refine generated segmentation maps. Although our CoAug module shares similar idea with co-segmentation methods, it aims to localize objects from images under the multi-category condition and further alleviates discriminative regions biased problem~\cite{hsu2018co}.

There are some other methods related to model interpretability but can also be applied to localization tasks. GradCAM~\cite{Selvaraju_2017_ICCV} combines gradient values and internal feature maps to produce CAM  without adding additional pooling layers. Chattopadhyay et al.~\cite{gradplus} further improve \cite{Selvaraju_2017_ICCV} by using a weighted combination of the positive partial derivatives of the feature maps in the last convolutional layer. These methods are usually engaged to propose new CAM that can interpret internal functions of various neural networks.  However, in this work, we focus on improving localization performance of CAM, which is a totally different purpose. Although we utilize original CAM~\cite{Zhou_2016_CVPR} in our method, the proposed modules can also be applied to different kinds of CAM as long as they share similar characteristics with the original one, i.e. highlighting target object parts as localization clues.

\begin{figure*}[t]
    \centering
    \includegraphics[width=0.9\linewidth]{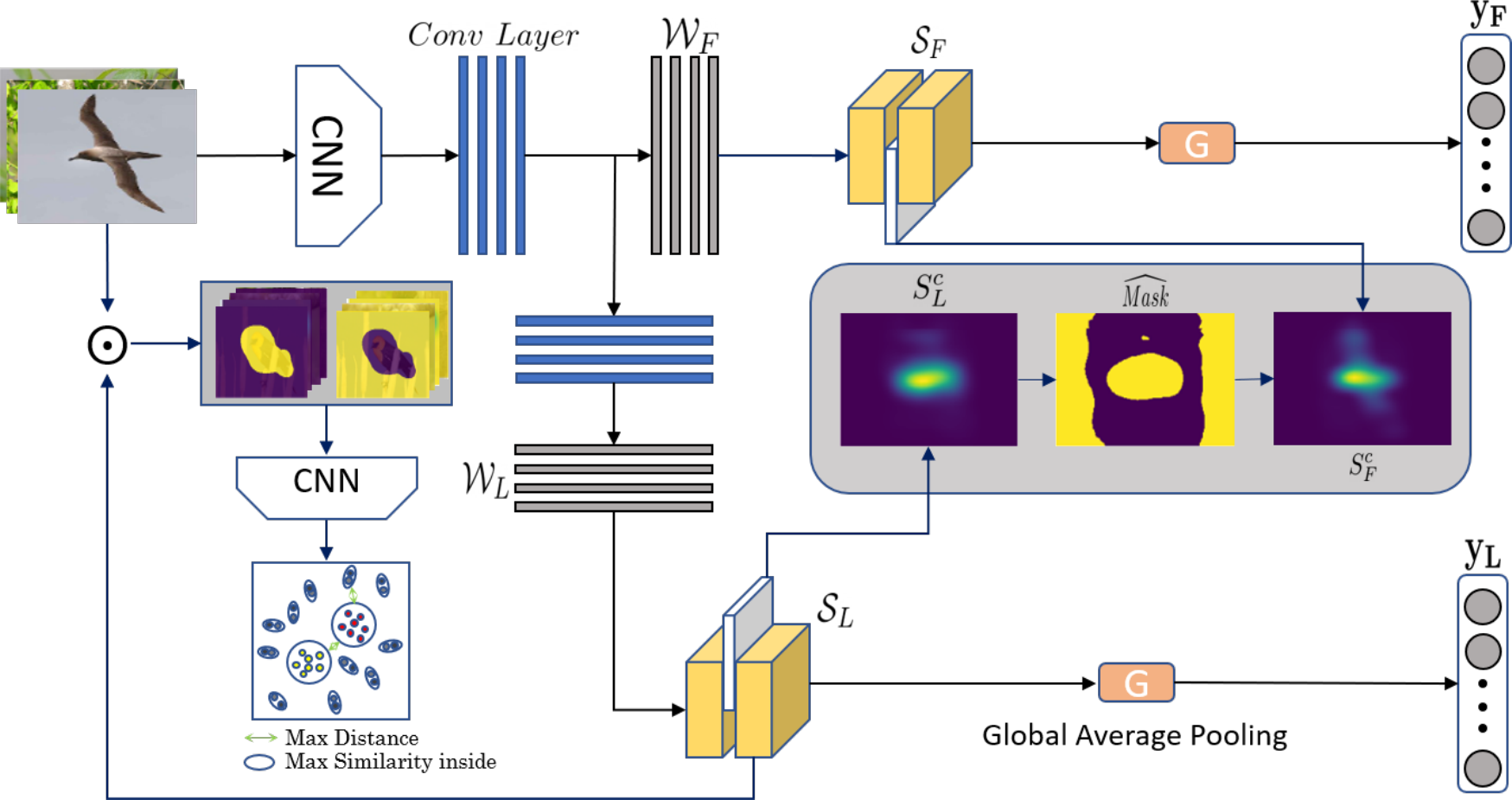}
    \caption{The overall structure of our CSoA network. For each input image, two different CAMs, $S_L$ and $S_F$, are generated from different classifiers and processed to logits for classification. Besides, the $c$-th slice of $S_L$ is then extracted and transformed to confidence segmentation mask by the ConfSeg module. The mask serves as additional supervisions by controlling the distance between $S_L^c$ and $S_F^c$. For samples in the same batch, they are first combined with foreground and background parts of $S_L^c$ separately. Finally all weighted samples are represented as 1-D vectors and play a metric-based learning process.}
    \label{fig:model}
\end{figure*}

\section{Method}
\label{sec:method}

In this section, we first review the seminal Class Activation Map (CAM)~\cite{Zhou_2016_CVPR}, then introduce our Confidence Segmentation (ConfSeg) module along with the Co-supervised Augmentation (CoAug) regularizer. An overview of our proposed method for the training phase is shown in Fig. \ref{fig:model}.

We first describe the weakly supervised object localization problem and the basic network proposed in \cite{Zhang_2018_CVPR} for generating CAM. Given a set of $N$ images, $\{I_n\}_{n=1}^N$ with $C$ categories, each image contains objects for only one category. Our goal is to classify each image and locate the corresponding objects with bounding boxes. In \cite{Zhang_2018_CVPR}, a Fully Convolutional Network (FCN) is proposed with a backbone $F$ consisting of $L$ layers, and a classifier $\mathcal{W_L}$. For an input image, the backbone network produces the feature map $M_l \in \mathbb{R} ^{K_l \times H_l \times W_l}$ after layer $l$ with $K_l$ channels. We denote $M_L \in \mathbb{R}^{K_L \times H_L \times W_L}$ as the last feature map from $F$. To generate CAM, the classifier $\mathcal{W_L}$ usually contains several convolutional layers to convert the number of channels from $K_L$ to $C$, i.e. the number of categories. Following that, a Global Average Pooling (GAP) layer is applied at each channel of $M_L$ to generate the class logit $\mathbf{y_L}=\{y_L^c\}_{c=1}^C$, which is then feed into the classification loss calculation. This process can be written as:
\begin{align}
S_L = \mathcal{W_L}(M_L) \;, 
\enspace
y_L^c = \frac{\sum_{i,j}(S_L^c)_{i,j}}{H_L \times W_L} \;, 
\enspace
\forall \; c \in \{1,\dots,C\} \;,
\end{align}
where $(S_L^c)_{i,j}$ refers to a certain pixel on the $c$-th channel of the feature map $S_L$. After training, the feature map $S_L^c$ corresponding to the predicted category is extracted. Then the largest connected region with positive values is segmented and finally processed to the bounding box prediction.

\subsection{Confidence Segmentation Module}
Though the basic framework is straightforward and efficient, it can only capture the most discriminative part of target objects. To address the problem, we propose the confidence segmentation (ConfSeg) module to generate a sample-specific confidence mask for CAM. The mask contains confidence score for each pixel and can segment regions with high confidence scores, including both foreground and background parts from CAM. With a high precision, the mask can serve as additional supervisions to guide other object detectors, encouraging them to explore more object-related regions.

To create another object detector that gets supervised, we extend one more CAM from internal feature maps inside the backbone network $F$, which can be denoted as $S_{F}$. Concretely, we first create a new classifier $\mathcal{W_F}$ that has the same structure with $\mathcal{W_L}$ and also goes through GAP layer to generate logits $\mathbf{y_F}$. Different from \cite{Xue_2019_ICCV} that builds several CAMs in multiple semantic levels, $\mathcal{W_L}$ and $\mathcal{W_F}$ share same categories for classification and have the same spatial size. As illustrated in \cite{Tang_2018_ECCV}, CNNs trained for object recognition have low-level vision features in early convolutional layers while more semantic features in top layers. Therefore, though $\mathcal{W_F}$ can be appended after any convolutional layer, it needs to localize objects precisely as well as achieving reasonable recognition performance. We will discuss the exact position for it in Section \ref{sec:exp}. 

With two different CAMs generated, the ConfSeg module connects them together. We first extract one slice from the feature map $S_L$, denoted as $S_L^c$ according to the ground truth index, or the predicted one in the inference time. Then we calculate the mean value of $S_L^c$, which is denoted as $\mu_1$. If the value of a pixel in $S_L^c$ is close to $\mu_1$, that means the corresponding position is ambiguous to be determined. In contrast, if a pixel has much larger or smaller value compared with $\mu_1$, it is very likely to be located on the target objects or background parts. Therefore, we can generate a confidence mask with each element calculated as the distance between each pixel and $\mu_1$. The process can be denoted as:
\begin{align}
Mask_{i,j} = \abs{(S_L^{c})_{i,j} - \mu_1},\; \text{where}\;\mu_1=\frac{\sum_{i,j}(S_L^{c})_{i,j}}{H_L \times W_L}.
\end{align}


After determining the confidence score for each pixel on $S_L^c$, the regions with high confidence are segmented from the mask by Eq.~\ref{equ:seg}. Instead of setting a fixed threshold for segmenting all image samples, we use a sample-dynamic threshold, denoted as $\mu_2$, for each image by taking the mean value of the mask. Therefore, the threshold for each sample is adaptively computed based on the corresponding confidence mask. If the score for one pixel is higher than $\mu_2$, we conclude that the pixel is very likely to have the correct value and vice versa. The equation can be formulated as:
%
\begin{equation}
\widehat{Mask_{i,j}} = 
\left\{
             \begin{array}{lr}
             1 ~~Mask_{i,j}>\mu_2\; &  \\
             0 ~~\text{otherwise}
             \end{array}
\right.
,
\label{equ:seg}
\end{equation}
\begin{equation}    
\text{where}~~\mu_2=\frac{\sum_{i,j}Mask_{i,j}}{H_L \times W_L}\;.
\end{equation}

With $S_{L}^c$ and the generated binary confidence mask $\widehat{Mask}$, we create a new supervision for $S_{F}^c$ by controlling the distance between each pixel in $S_{L}^c$ and $S_{F}^c$. For the positions that their corresponding values are $1$ in $\widehat{Mask}$, we encourage $S_{F}^c$ to be similar with $S_{L}^c$, which means $S_{F}^c$ should follow the decisions made by $S_{L}^c$ if they are confident enough. For other positions, we allow $S_{F}^c$ to be different from $S_{L}^c$ so that it can refine the object boundaries when the decisions are made with low confidence. With such an adversarial strategy, we do not need to worry if additional explorations by $S_{F}^c$ for object related regions may reach background parts because the confidence mask sets solid restriction for the background part in CAM. We formulate the process as the following loss: 
\begin{align}
\mathcal{L}_{inner} = \sum_{i,j}|(S_F^c)_{i,j}-(S_L^c)_{i,j}|\odot\widehat{Mask_{i,j}}\;.
\end{align}
Finally, the total loss function with the ConfSeg module is: 
\begin{align}
\mathcal{L_C} = \mathcal{L}_{cls}+\alpha \mathcal{L}_{inner}\;,
\end{align}
where $\alpha$ is a factor ranging within $[0, 1]$ that increases along the training epoch to avoid unstable prediction from $S_{L}$ at the initial training process. $\mathcal{L}_{cls}$ denotes the cross entropy loss for both $\mathbf{y_L}$ and $\mathbf{y_F}$ with same ground-truth categories.

\noindent \textbf{The relation to Zhang et al.~\cite{Zhang_2018_ECCV}.} \quad By further formulating our proposed ConfSeg module, we show that it is a generalized version of Zhang et al.~\cite{Zhang_2018_ECCV}. The latter sets pre-fixed thresholds of foreground and background for all image samples before the training process. Our method can also represent their thresholds through simple transformation, which can be denoted as:
\begin{equation}
\begin{split}
\xi_1 &= \mu_1+\mu_2\;,\\
\xi_2 &= \mu_1-\mu_2\;,
\end{split}
\end{equation}
where $\xi_1$ and $\xi_2$ are thresholds for foreground and background, respectively. Therefore, the ConfSeg module can achieve the same function as \cite{Zhang_2018_ECCV} but is versatile with sample-level adaption for the segmentation of CAM without introducing additional parameters. 

\subsection{Co-supervised Augmentation Module}


For the fully supervised localization task, the ground-truth box annotations are always utilized to guide the generation of object proposals. However, in the setting of WSOL task, only image-level supervisions are available, which leads to severe bias of recognition models that tends to localize the most discriminative region rather than entire objects. To address the problem, we further introduce a plug-in metric-based module to regulate CAM with feature-level supervisions, since the comparison between different samples is capable of preserving more visual features.

Our approach is inspired by metric learning methods~\cite{koch2015siamese} that embed images into representation vectors and leverage distance as metrics to estimate their correlations. Therefore, in CoAug module, we consider two kinds of relationships: 1) foreground and background part that should both represent distinct features; 2) foreground objects of different samples in the batch level.





Before discussing the details about metric-based processes, we first segment out predicted foreground and background regions of input images according to generated CAM. For the reason that CAM is able to highlight foreground object region of target category, we multiply the slide of CAM according to ground-truth index with raw input images to represent corresponding object and then embed the object into feature vector $F_m$. Similarly, we can also obtain background vector $B_m$ denoted as:
\begin{equation}
\begin{split}
F_m &= \mathcal{W_E}(S_l^{c_m} \odot I_m)\;,\\
B_m &= \mathcal{W_E}((1 - S_l^{c_m}) \odot I_m)\;,
\end{split}
\label{equ:embed}
\end{equation}
where $\mathcal{W_E}$ indicates the embedding network, $\odot$ represents pixel-wise multiplication, $S_l^{c_m}$ refers to the localization map from the $c_m$ channel of $S_l$, and $c_m$ is the $c$-th category of the $m$-th image. Please note that $I_m$ can be either the intermediate feature map generated by a CNN or directly the raw $m$-th image. 

Then the Relation.1 can be measured as the distance between $F_m$ and $B_m$ as :
\begin{equation}
D^{cam}_m = \left\lVert F_m - B_m\right\rVert_2\;.
\label{equ:cam_dis}
\end{equation}
in which we expect $D^{cam}_m$ to be large. Additionally, we utilize the background part of the confidence mask introduced in the previous section to augment the ability of CAM to avoid mis-classifying foreground part as Eq.~\ref{equ:bg_dis}: 
\begin{equation}
\begin{split}
D^{back}_m &= \left\lVert B_m - Mask_B^m\right\rVert_2\;,\\
\text{where } Mask_B^m &= \mathcal{W_E}((1-S_l^{C_m}) \odot \widehat{Mask} \odot I_m)\;.
\end{split}
\label{equ:bg_dis}
\end{equation}
Specifically, we obtain $Mask^m_B$ by first multiplying $\widehat{Mask}$ with $I_m$ to extract confident parts in the image, and then incorporating it with $1-S_l^{c_m}$ to represent the background content.

Apart from comparing foreground and background regions of a single image, we also consider the relationship among multiple samples, denoted as Relation.2 above. We calculate the distance between foreground vectors of different input samples as:
\begin{align}
D_{m,n} &= \left\lVert F_m-F_n \right\rVert_2.
\label{equ:img_dis}
\end{align}
When $F_m$ and $F_n$ belong to the same category, they are supposed to share similar representations and $D_{m,n}$ should be small. In this case we change $D_{m,n}$ to $D^{same}_{m,n}$. For other cases, where the categories of $F_m$ and  $F_n$ are different, we convert $D_{m,n}$ to $D^{diff}_{m,n}$ and expect it to be large.

With the definition of the four distances for regulation, we define the following loss function to have images supervising each other:
\begin{equation}
\begin{split}
\mathcal{L}_D^{same} &= \sum_{\{m,n|cls(m)\neq cls(n)\}} \frac{\gamma \cdot (D^{back}_m+D^{back}_n)}{\delta \cdot D^{diff}_{m,n}+\frac{1}{2}(D^{cam}_m+D^{cam}_n)},\\
\mathcal{L}_D^{diff} &= \sum_{\{m,n|cls(m)=cls(n)\}} \frac{\gamma \cdot (D^{back}_m+D^{back}_n) + D^{same}_{m,n}}{\frac{1}{2}(D^{cam}_m+D^{cam}_n)},\\
\mathcal{L}_D &= \mathcal{L}_D^{same} + \mathcal{L}_D^{diff},
\end{split}
\label{equ:dis_loss}
\end{equation}
where $\gamma$ and $\delta$ in the equation are two factors that controls the global scale of $\mathcal{L_D}$, while $cls(\cdot)$ refers to category. 

For the training time, we combine $\mathcal{L_C}$ and $\mathcal{L_D}$ together. During the inference time, we remove both ConfSeg and CoAug module and only keep the two generated CAMs. We first segment the target object parts following the instruction in \cite{Zhou_2016_CVPR}. In details, we extract max values $S_F^{max}$ and $S_L^{max}$ from $S_F$ and $S_L$ respectively. Then we create binary localization maps by Eq. \ref{equ:bi_cam} denoted as:
\begin{equation}
\widehat{S_{F/L}} = 
\left\{
             \begin{array}{lr}
             1 ~~S_{F/L}^{i,j}>\theta \cdot S_{F/L}^{max} \;,\\
             0 ~~\text{otherwise} \;,
             \end{array}
\right.
\label{equ:bi_cam}
\end{equation}
where $\theta$ is a pre-defined parameter for segmentation. Finally, we combine the two localization maps with the pixel value as $1$ if either pixel value on $\widehat{S_{F}}$ or $\widehat{S_{L}}$ is $1$. Otherwise, the pixel value is set to $0$ since neither of two CAMs consider it belonging to foreground object parts.

\section{Experiment}
\label{sec:exp}

\subsection{Implementation Details}
Following the configuration of previous methods~\cite{Zhang_2018_ECCV,Xue_2019_ICCV}, our proposed modules are integrated with the commonly used CNNs including VGGnet~\cite{simonyan2014a} and GoogLeNet~\cite{SzegedyLJSRAEVR14}. We construct the same structure for both classifiers $\mathcal{W_F}$ and $\mathcal{W_L}$. The structure consists of two convolutional layers with kernel size $3 \times 3$, stride $1$, pad $1$ with
$1024$ units, and a convolutional layer of size $1 \times 1$, stride $1$ with $1000$ units ($200$ units for CUB-200-2011). For GoogLeNet, we remove the covolutional layers after $Mixed\_6e$ to increase the resolution of the final output. The two classifiers are appended after the layer $Mixed\_6b$ and $Mixed\_6e$ respectively. For VGGNet, we remove the final linear layer and append two classifiers after the fourth and final convolutional block. We will discuss the performance of our model in Section \ref{sec:exp} with different positions applied for appending $\mathcal{W_F}$.

For the CoAug module, we apply Alexnet~\cite{Alexnet} as the feature extractor for both estimated foreground and background regions of input samples. The module utilizes $S_L$ as the only CAM for segmentation. The batch size is set to $48$ with at most $12$ categories for each batch. All the networks are fine-tuned with the pre-trained weights of ILSVRC2016~\cite{5206848}. We train the model with an initial learning rate of $0.001$ and decay of $0.95$ each epoch. The optimizer is SGD with $0.9$ momentum and $5\times 10^{-4}$ weight decay. For classification result, we follow the instructions in~\cite{Zhou_2016_CVPR}, which further averages the scores from the softmax layer with $10$ crops.

\begin{table}[t]
    \centering
    \caption{Effect of our individual modules on CUB-200-2011}
    \vspace{2mm}
    \begin{tabular}{ l | c c | c c }
    \toprule
        \multirow{2}{*}{\textbf{Methods}} & \multicolumn{2}{c|}{\textbf{Loc. Error}} & \multicolumn{2}{c}{\textbf{Class. Error}} \\\cline{2-3}\cline{4-5}
         & Top-1 & Top-5 & Top-1 & Top-5\\
        \midrule
        VGGnet-DANet~\cite{Xue_2019_ICCV} & 47.48  & 38.04 & 24.12 & 7.73\\ 
        VGGnet-base & 53.42  & 45.85 & 24.73 & 8.96\\ 
        VGGnet-ConfSeg & 39.02 & 27.17 & 23.14 & 6.94 \\ 
        VGGnet-CoAug & 40.78 & 29.36 & 23.06 & 6.77 \\
        VGGnet-CSoA &  \textbf{37.69} & \textbf{26.49} & \textbf{21.41} & \textbf{5.94}\\    
    \bottomrule
    \end{tabular}
    \label{tab:ab_module}
\end{table}

\subsection{Experiment Setup}
\textbf{Dataset and Evaluation} \quad To draw a fair comparison, we test our model on ILSVRC2016~\cite{5206848} validation set and CUB-200-2011~\cite{WahCUB_200_2011} test set, which are two most widely-used benchmarks for WSOL.
The ILSVRC dataset has a training set containing more than 1.2 million images of 1,000 categories and a validation set of 50,000 images. In CUB-200-2011, there are totally 11,788 bird images of 200 classes, among which 5,994 images are for training and 5,794 for testing. We leverage the localization metric suggested by~\cite{Russakovsky2015}.
Specifically, the bounding box of an image is correctly predicted if: 1) the model predicts the right image label; 2) more than 50\% Intersection-over-Union (IoU) is observed in the overlapped area between predicted bounding boxes and ground truth boxes. For more details, please refer to \cite{Russakovsky2015}. We also note that a very recent work by Choe et al.~\cite{Choe_2020_CVPR} proposes a new set of evaluation metrics providing new perspectives to the evaluation. However, we still use the traditional evaluation metrics, i.e. localization and classification errors, in this work for their feasibility to benchmark with a wide spectrum of existing methods.

\begin{table}[t]
    \centering
    \caption{Effect of positions to insert additional classifier on CUB-200-2011. The number after our model indicates the order of convolutional block in VGGnet}
    \vspace{2mm}
    \begin{tabular}{ l | c c | c c }
    \toprule
        \multirow{2}{*}{\textbf{Methods}} & \multicolumn{2}{c|}{\textbf{Loc. Error}} & \multicolumn{2}{c}{\textbf{Class. Error}} \\\cline{2-3}\cline{4-5}
         & Top-1 & Top-5 & Top-1 & Top-5\\
        \midrule
        CSoA-3-5 & 52.76 & 41.78 & 26.98 & 8.58\\ 
        CSoA-4-5 & \textbf{37.69} & \textbf{26.49} & \textbf{21.41} & \textbf{5.94} \\
        CSoA-5-5 & 55.61 & 44.72 & 28.60 &9.23\\    
    \bottomrule
    \end{tabular}
    \label{tab:ab_pos}
\end{table}

\begin{table}
    \centering
    \caption{Localization error with different thresholds for segmentation}
    \vspace{2mm}
    \begin{tabular}{c|c|c}
    \toprule
    Thresholds & Top-1 Error & Top-5 Error \\\hline
    $0.2$ & 39.13 & 27.58\\\hline
    $0.3$ & \textbf{37.69} & \textbf{26.49}\\\hline
    $0.4$ & 38.82 & 27.11\\
    \bottomrule
    \end{tabular}
    \label{tab:thre}
    \vspace{-0.2in}
\end{table}

\subsection{Ablation Studies}
We make some ablation studies on CUB-200-2011 using VGGnet to evaluate the effects of our individual modules. Besides, we discuss the functions of some hyper-parameters related to the network structure and the inference process.

\noindent \textbf{Effect of ConfSeg and CoAug:} For the fair comparison, we first construct a baseline network according to~\cite{Zhang_2018_CVPR} which consists of VGGnet as the backbone and a classifier with the same structure as $\mathcal{W_F}$. As shown in Table~\ref{tab:ab_module}, the performance of the network with only ConfSeg module reduces the top-1/top-5 \textit{loc. err} by over $14\%/18\%$ compared with our baseline model, and $8\%/10\%$ compared
with~\cite{Xue_2019_ICCV} respectively. It demonstrates that two interacted CAMs can significantly decrease the localization error since the value of each pixel on the final localization map is double confirmed by both classifiers.

For our model with CoAug module only, the localization result is also much better than the baseline and \cite{Xue_2019_ICCV}. Please note that the CoAug module do not have any modification inside the model structure. It only regulates generated CAMs from batch-level, serving as a clustering method among samples with various categories. Therefore, the CoAug module is general enough to be applied to any kind of network as long as the network can generate CAMs-like feature maps.

\begin{table}[t]
    \centering
    \caption{Performance comparison on the CUB-200-2011 test set. The method with star apply a novel non-local approach on the backbone to boost the performance.}
    \vspace{2mm}
    \begin{tabular}{l| c | c | c | c}
    \toprule
        \multirow{2}{*}{\textbf{Methods}} & \multicolumn{2}{c|}{\textbf{Loc. Error}} & \multicolumn{2}{c}{\textbf{Class. Error}} \\\cline{2-3}\cline{4-5}
         & Top-1 & Top-5 & Top-1 & Top-5\\\hline
        GoogLeNet-GAP~\cite{Zhou_2016_CVPR} & 58.94 & 49.34 & 35.0 & 13.2\\
        GoogLeNet-SPG~\cite{Zhang_2018_ECCV} & 53.36 & 42.28 & - & -\\
        GoogLeNet-ADL~\cite{Choe_2019_CVPR} & 46.96 & - & 25.4 & -\\
        GoogLeNet-DANet~\cite{Xue_2019_ICCV} & 50.55 & 39.94 & 28.8 & 9.4\\\hline
        GoogLeNet-Ours & \textbf{46.06} & \textbf{34.36} & \textbf{23.9} & \textbf{6.4} \\\hline
        VGGnet-GAP~\cite{Zhou_2016_CVPR} & 55.85 & 47.84 & 23.4 & 7.5\\
        VGGnet-ACoL~\cite{Zhang_2018_CVPR}& 54.08 & 43.49 & 28.1 & -\\
        VGGnet-SPG~\cite{Zhang_2018_ECCV}& 51.07 & 42.15 & 24.5 & 7.9\\
        VGGnet-ADL~\cite{Choe_2019_CVPR} & 47.64 & - & 34.7 & -\\
        VGGnet-DANet~\cite{Xue_2019_ICCV} & 47.78 & 38.04 & 24.6 & 7.7\\
        NL-CCAM*~\cite{combine_wacv20}& 47.60 & 34.97 & 26.6 & -\\\hline
        VGGnet-Ours & \textbf{37.69} & \textbf{26.49} & \textbf{21.4} & \textbf{5.9} \\    
    \bottomrule
    \end{tabular}
    \label{tab:cub_loc}
    \vspace{-0.2in}
\end{table}

Finally, our model that combines the ConfSeg and CoAug module can outperform all previous ones on both localization and classification tasks. Especially, the \textit{cls. err} is reduced by about $2\%$ on both top-1 and top-5 results, which is valuable since lots of methods~\cite{Zhou_2016_CVPR,Choe_2019_CVPR,Xue_2019_ICCV} cannot keep the classification performance when trying to improve the localization ability. We mainly attribute it to the double classifiers that refine the bottom layers of our network. Besides, the CoAug module also regulates the network, enforcing it to recognize different parts of the target objects rather than only the most discriminative region.

\noindent \textbf{Position for $\mathcal{W_F}$:} For applying ConfSeg module, we extend one more classifier $\mathcal{W_F}$ from the backbone network to obtain additional CAM for interaction. Table~\ref{tab:ab_pos} shows the results for inserting $\mathcal{W_F}$ after different convolutional blocks. We can obtain the best result when inserting $\mathcal{W_F}$ after the fourth block in VGGNet, i.e. the block right before the final block. In such a configuration, $\mathcal{W_F}$ can do the classification task with features in high semantic level and also produce CAM with different receptive fields. It encourages the effective interaction between two CAMs, which makes their decisions more complementary on ambiguous regions.

\noindent \textbf{Thresholds for Binary Mask:} During the inference, we need one threshold $\theta$ to extract foreground regions from two CAMs and then combine them together. To inspect their influence for the localization result, we test different thresholds for our model in Table \ref{tab:thre}. We can see our model achieves the best result with $\theta = 0.2$. However, $\theta$ is still a pre-defined parameter for extracting the final target object. How to remove it or how to make it learnable may be a future work for us to explore.

\begin{figure*}[t]
    \centering
    \includegraphics[width=1\textwidth]{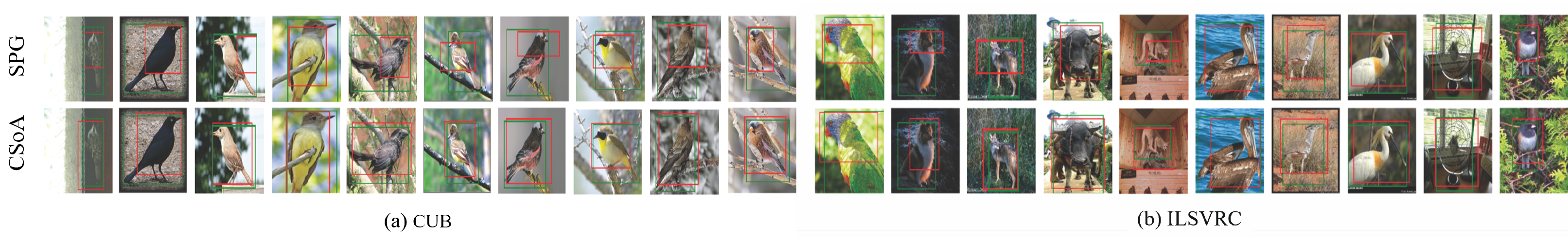}
    \caption{Compare localization examples between SPG and CSoA. All visual results from SPG are generated by strictly following author-released code.}
    \label{fig:compare}
\end{figure*}

\begin{figure*}[t]
    \centering
    \includegraphics[width=1.0\textwidth]{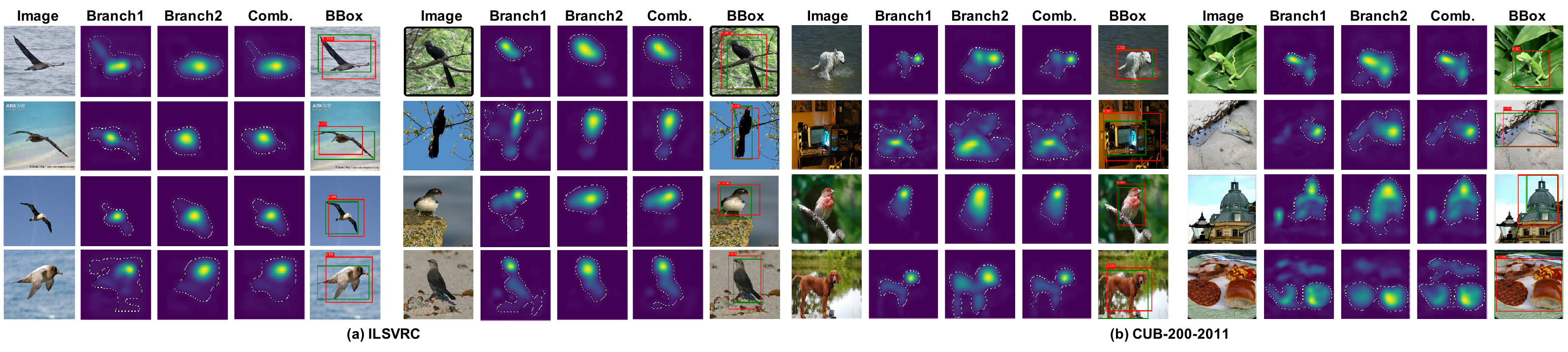}
    \caption{Output visual examples of CSoA. For each dataset, the first three rows show successful results while the last row provides two examples that fail to connect detected parts together.}
    \label{fig:visual}
    \vspace{-0.1in}
\end{figure*}

\subsection{Comparison with the state-of-the-arts}

\begin{table}[t]
    \centering
    \caption{Performance comparison on the ILSVRC test set}
    \vspace{2mm}
    \begin{tabular}{l| c | c | c | c}
    \toprule
        \multirow{2}{*}{\textbf{Methods}} & \multicolumn{2}{c|}{\textbf{Loc. Error}} & \multicolumn{2}{c}{\textbf{Class. Error}} \\\cline{2-3}\cline{4-5}
         & Top-1 & Top-5 & Top-1 & Top-5\\
        \midrule
        VGGnet-BP~\cite{simonyan2014a} & 61.12 & 51.46 & - & -\\
        VGGnet-GAP~\cite{Zhou_2016_CVPR} & 57.20 & 45.14 & 33.4 & 12.2\\
        VGGnet-Grad~\cite{Selvaraju_2017_ICCV} & 56.51 & 46.41 & 30.4 & 10.9\\
        VGGnet-ACoL~\cite{Zhang_2018_CVPR}& 54.17 & 40.57 & 32.5 & 12.0\\
        VGGnet-ADL~\cite{Choe_2019_CVPR} & 55.08 & - & 39.3 & -\\
        VGGnet-CCAM~\cite{combine_wacv20}& 51.78 & 40.64 & 33.4 & -\\
        NL-CCAM*~\cite{combine_wacv20}& 49.83 & 39.31 & 27.7 & -\\
        \midrule
        GoogLeNet-BP~\cite{simonyan2014a} & 61.31 & 50.55 & - & -\\
        GoogLeNet-GAP~\cite{Zhou_2016_CVPR} & 56.40 & 43.00 & 35.0 & 13.2\\
        GoogLeNet-ACoL~\cite{Zhang_2018_CVPR}& 53.28 & 42.58 & 29.0 & 11.8\\
        GoogLeNet-SPG~\cite{Zhang_2018_ECCV} & 51.40 & 40.00 & - & -\\
        GoogLeNet-ADL~\cite{Choe_2019_CVPR} & 51.29 & - & \textbf{27.2} & -\\
        GoogLeNet-DANet~\cite{Xue_2019_ICCV} & 52.47 & 41.72 & 27.5 & \textbf{8.6}\\
        \hline
        GoogLeNet-CSoA &  \textbf{48.81} & \textbf{37.46} & 28.1 & 9.8\\    
    \bottomrule
    \end{tabular}
    \label{tab:ILSVRC_res}
    \vspace{-0.2in}
\end{table}

We compare our CSoA with state-of-the-art methods on CUB-200-2011 test set and ILSVRC validation set.

As shown in Table~\ref{tab:cub_loc}, on CUB-200-2011 test set, with VGGnet as the backbone network, our method outperforms all others by more than $10\%$ on both Top-1 and Top-5 \textit{loc. err}. It demonstrates the powerful localization ability of our proposed modules with the simple backbone structure. Besides, the classification results of our model are much better than other WSOL methods, which indicates that our proposed method has little negative impact on the recognition performance. This property is important for some real applications, e.g. surveillance cameras that prefer to classify objects correctly and also estimate their positions.

We also evaluate our model with GoogLeNet. Though not as good as VGGnet, our model also becomes the state-of-the-art compared with others. We infer that the smaller gap between GoogLeNet-CSoA and others is because of combination of various operations together for input features in each layer, e.g. \textit{pooling}, $3\times 3$ and $1\times 1$
convolution kernels. It reduces the difference in receptive field between layers, which makes it challenging for multiple classifiers to explore in various levels. The results with Resnet~\cite{resnet} backbone that is only reported in \cite{Choe_2019_CVPR} with $37.71\%$ Top-1 \textit{loc. err} has the similar problem since the residual link connects different blocks to reduce the receptive differences. 

Table~\ref{tab:ILSVRC_res} shows both classification and localization results on ILSVRC validation set with GoogLeNet. For the localization, our result outperforms all others with the same backbone by over $2\%$ on Top-1 \textit{loc. err}. Besides, our model also achieve better performance compared to the methods with VGGnet. Especially, the NL-CCAM proposed in \cite{combine_wacv20} uses a novel non-local backbone to improve the localization performance, which can be also integrated with our method.

To further demonstrate the localization ability of our model and make a full comparison with other methods, we use ground-truth classification labels for ILSVRC validation set and only evaluate localization performance serving as an ``upper-bound''~\cite{Zhang_2018_ECCV}. Denoted as GT-Known \textit{loc. err} in Table~\ref{tab:ILSVRC_loc_star}, our result is still better than others, which indicates an advantage in terms of the pure localization.

\begin{table}
    \centering
    \caption{GT-Known localization results for ILSVRC validation set}
    \vspace{2mm}
    \begin{tabular}{l|c}
    \toprule
        \textbf{Methods} & \textbf{Top-1 loc. err} \\\hline
        AlexNet-GAP~\cite{Zhou_2016_CVPR} & 45.01\\
        AlexNet-HaS~\cite{Singh_2017_ICCV} & 41.26\\
        GoogLeNet-GAP~\cite{Zhou_2016_CVPR} & 41.34\\
        GoogLeNet-HaS~\cite{Singh_2017_ICCV} & 39.43\\
        VGGnet-ACoL~\cite{Zhang_2018_CVPR}& 37.04 \\
        SPG~\cite{Zhang_2018_ECCV} & 35.31\\
        ADL~\cite{Choe_2019_CVPR} & 34.59\\\hline
        GoogLeNet-CSoA & \textbf{33.80}\\
        \bottomrule
    \end{tabular}
    \label{tab:ILSVRC_loc_star}
    \vspace{-0.2in}
\end{table}

Figure~\ref{fig:compare} visualizes the comparison result between our localization results with SPG~\cite{Zhang_2018_ECCV} since it also considers both foreground and background parts. In most situations, our method can generate more precise bounding boxes than SPG, which demonstrates that the sample-specific segmentation method can achieve better results than using the same pre-fixed threstholds for all images. We will provide more convincing visual examples in the \textbf{appendix.}  

In addition, in Fig.~\ref{fig:visual}, we visualize the localization maps from CAM at both classifiers, the combined CAM and the final bounding box result of our proposed method on both ILSVRC and CUB-200-2011. The areas inside dashed lines for each CAM indicate the segmented regions for the final result. We can observe that in most cases, the combination of two CAMs has a more stable localization result than any single CAM. It collects the final pixels that are determined by both CAMs and removes ambiguous pixels.

\section{Conclusion}
We propose CSoA, a novel method for WSOL task. The method consists of two modules that refine the traditional convolutional networks to improve their localization performance without the sacrifice of recognition ability. During learning, the ConfSeg module encourage two classifiers inside the network to generate more precise and complete CAM. In addition, the CoAug module regulate CAM from different samples based on metric approaches in batch level. Our final model outperforms all previous approaches on two public benchmarks. It becomes the new state-of-the-art and provides fresh insights for tackling the WSOL problem.


\section*{Acknowledgments}
This research was funded in part by the Center of Excellence in Data Science, an Empire State Development-Designated Center of Excellence.


{\small
\bibliographystyle{ieee_fullname}
\bibliography{egbib}
}

\end{document}